\documentclass{article}
\usepackage[english]{babel}
\usepackage[utf8]{inputenc}
\usepackage[letterpaper,top=2cm,bottom=2.5cm,left=3cm,right=3cm,marginparwidth=1.75cm]{geometry}

\usepackage{float}
\usepackage{paralist}      % gives enumerate*, itemize*, compact lists
\usepackage{enumitem}      % nicer list customisation (optional but handy)
\usepackage{amsmath}
\usepackage{booktabs}      % \toprule, \midrule, \bottomrule in tables
\usepackage{siunitx}       % \SI{...}{...} and \num{...}
\usepackage{multirow}
\usepackage{amssymb,amsfonts} % gives \mathbb, \mathcal, etc.
\usepackage{graphicx}
\usepackage[table,xcdraw]{xcolor}
\usepackage{caption}
\usepackage{subcaption}
\usepackage[colorlinks=true, allcolors=blue]{hyperref}

\usepackage{multicol}
\usepackage{array}
\usepackage{xcolor}
\usepackage{tabularx}
\usepackage{rotating}
\usepackage[T1]{fontenc}
\newcolumntype{C}[1]{>{\centering\arraybackslash}p{#1}}
\newcolumntype{L}[1]{>{\raggedright\arraybackslash}p{#1}}
\newcolumntype{R}[1]{>{\raggedleft\arraybackslash}p{#1}}

\newcommand{\best}[1]{\textbf{#1}}

\providecommand{\keywords}[1]{
\small\textbf{\textit{Keywords---}} #1
}

\title{TCAR-Gen: Temporal Graph Retrieval with Evidence Fusion for Knowledge-Grounded Generation}
\usepackage{authblk}

\begin{document}

%\author{Sidra Nasir \\ rnsidra5@gmail.com}

\author{
   Sidra~Nasir$^{1}$,
    Muhammad~Noman~Zahid$^{2}$,
     Rizwan~Ahmed~Khan$^{3*}$,\\ 
    \textit{$^{1}$Dipartimento di Informatica, Università di Verona, Italy}\\
    \textit{$^{2}$ School of Advanced Studies, University of Camerino, Italy}\\
    \textit{$^{3}$ Department of Computer Science, School of Mathematics and Computer Science, Institute of Business Administration (IBA), Karachi, Pakistan}\\
    
    *Corresponding author: Rizwan Ahmed Khan (email: rizwankhan@iba.edu.pk)  
}% <-this % stops a space

\date{} % This will remove the date.

\maketitle
\begin{abstract}
Retrieval-augmented generation systems struggle with temporal reasoning and evidence fusion when answering complex questions over historical criminal case narratives. Existing approaches either retrieve independently of query semantics or fail to integrate multiple evidence sources coherently. We propose Temporal Context Augmented Retrieval Generation (TCAR-Gen), a framework that combines query-conditioned graph neural networks, temporal evidence fusion, and chain-of-trees reasoning to ground answer generation in retrieved evidence. On the Victorian Crime Diaries benchmark, TCAR-Gen achieves 0.3738 Recall@5, outperforming Vanilla RAG, Temporal RAG, GraphRAG-C, and GraphRAG-T across seven query types including multi-hop reasoning and counterfactual questions. Ablation studies reveal that the context graph, temporal penalty mechanism, and query conditioning are critical components. Cross-model evaluation across five language model (GPT-OSS 20B to TinyLlama 1.1B) demonstrates that TCAR-Gen maintains robust retrieval coverage at smaller model scales, though generation quality degrades substantially with reduced model capacity. Our work shows that explicit temporal modelling and multi-branch evidence fusion are essential for faithful, reasoning-intensive question answering over knowledge-grounded corpora.
\end{abstract}

\section{Introduction}\label{sec1}

In recent years, Large Language Models (LLMs) have advanced natural language processing and demonstrated strong performance across a wide range of tasks, including summarization, machine translation, and question answering. Their ability to generalize across tasks with limited task-specific supervision has made them suitable for deployment in diverse application domains~\cite{qin2023chatgpt}. However, reliability remains a central concern, particularly in applications where responses must be factually grounded, contextually precise, and open to verification. A key limitation of LLMs is the generation of outputs that are linguistically fluent but factually unsupported or logically inconsistent with available evidence. This behaviour, commonly referred to as hallucination, becomes critical in knowledge-intensive and domain-specific tasks, where correct answers depend on access to external information and on the ability to organise that information coherently~\cite{peng2023check}. In such scenarios, knowledge stored within model parameters is often insufficient, especially when tasks require linking multiple pieces of evidence, resolving dependencies among entities or events, or reasoning over temporally distributed information.

Retrieval-augmented generation addresses this limitation by supplying LLMs with external textual evidence at inference time~\cite{borgeaud2022improving, ram2023incontext}. Although this approach improves factual grounding, most retrieval pipelines rely on similarity-based matching between queries and isolated text chunks. This design favours local semantic overlap over broader structural relationships, which results in evidence that is incomplete, fragmented, or poorly organised for downstream reasoning, particularly in multi-hop scenarios involving interconnected facts. Contextual retrieval methods extend this approach by enriching document chunks with surrounding document-level information before indexing or ranking. This improves interpretability by situating each chunk within its local context. Even so, this strategy remains limited when meaning depends on explicit relationships among entities, events, and temporally ordered evidence rather than on textual proximity alone. Knowledge Graphs (KGs) provide a natural representation for such structure by organising information through entities and relations. Graph Neural Networks (GNNs) complement this representation by learning over relational data through neighbourhood aggregation and higher-order dependency modelling~\cite{lin2019kagnet, feng2020mhgrn, li2023graphllm}. These developments indicate that effective question answering in complex settings requires preserving the structural and temporal organisation of evidence rather than relying solely on text-based retrieval.

The present work addresses this requirement by formulating retrieval and generation as a unified reasoning process. Instead of treating retrieval as a separate preprocessing step, the proposed framework integrates contextual, relational, and temporal signals into a single inference pipeline. This design combines contextual chunk enrichment, query-conditioned graph construction, temporal encoding, and multi-branch reasoning to support evidence selection and answer generation in a coordinated manner. The contributions of this work are threefold. First, it proposes a context-aware retrieval framework that integrates document-level enrichment with query-conditioned graph construction and temporal modelling. Second, it introduces a reasoning pipeline that combines structured evidence retrieval with multi-branch inference grounded in relational and temporal constraints. Third, it provides a comprehensive empirical evaluation that examines retrieval effectiveness, generation faithfulness, and the contribution of individual components across multiple model scales.

The remainder of this paper is organised as follows. Section~\ref{sec:rl} reviews related work on retrieval-augmented generation, graph-based reasoning, temporal modelling, and multi-step inference. The methodology section \ref{sec:methodology} introduces the proposed framework and its components, including context graph construction, hybrid retrieval, and reasoning mechanisms. The experimental section \ref{sec:experiments} describes the dataset, evaluation protocol, and baselines. In section \ref{sec:results} results are then presented and analysed through comparative evaluation, ablation studies, and scaling behaviour. The paper concludes with a discussion of findings, limitations, and directions for future research.

\section{Related Work} \label{sec:rl}

Retrieval-augmented generation (RAG) improves the factual grounding of
large language models by combining parametric generation with external
non-parametric knowledge. Early work showed that retrieval at inference
time improves performance on knowledge-intensive tasks by reducing
dependence on information stored only in model
parameters~\cite{lewis2020rag}. Later studies showed that retrieval
quality remains central to downstream generation, since more relevant and
informative context improves both faithfulness and
accuracy~\cite{ram2023incontext, borgeaud2022improving, hu2025context}. Recent methods have moved retrieval closer to the decoding process itself, which allows generation to use retrieved evidence during multi-step
reasoning~\cite{feng2025retrieval}. This line of work establishes
retrieval quality as a core factor in reliable generation.

Most RAG systems still rely on semantic similarity between a query and
isolated text chunks. This strategy retrieves locally relevant passages
effectively, but it often fails to preserve document-level context and
does not handle multi-hop reasoning well when evidence is distributed
across related passages. Contextualised retrieval methods partly address
this issue by enriching chunks with surrounding document information.
Even so, these methods remain largely text-centric and do not explicitly
represent relationships between pieces of evidence. The limitation is
therefore not only retrieval coverage but also the absence of structured
evidence modelling.

Graph-based retrieval addresses this limitation by representing evidence
through explicit relations. Early systems such as
GRAFT-Net~\cite{sun2018graftnet} and PullNet~\cite{sun2019pullnet}
showed that question-specific subgraphs constructed from text and
structured knowledge improve multi-hop reasoning. These studies also
showed that retrieval and reasoning are more effective when treated as
joint and iterative processes rather than as separate stages. Later work
extended this idea to large language models. KGLLM links entity mentions
in a query to an external knowledge graph, extracts surrounding
subgraphs, linearises them into natural language, and uses this evidence
to ground generation and re-rank candidate outputs for factual
consistency~\cite{yang2024give}. This direction shows that grounding
generation in verified external evidence reduces hallucination more
effectively than reliance on parametric memory alone.

GraphRAG~\cite{edge2024graphrag} organises retrieved evidence into graph
structures to support coherent synthesis over large corpora, whereas
G-Retriever~\cite{he2024gretriever} performs retrieval directly over
textual graphs. Structured graph traversal has also improved factual
grounding in systems that decompose complex queries into sub-questions
before reasoning~\cite{linders2025knowledge}. RDPG extends this approach
through adaptive path generation, where an LLM iteratively explores a
knowledge graph, revises candidate paths, and integrates recovered paths
into a chain-of-thought prompt for final answer
generation~\cite{ding2025rdpg}. Similar graph-based retrieval has also
improved reliability and interpretability in domain-specific tasks such
as autonomous driving~\cite{hussien2025rag}. Multi-level graph
representations further improve reasoning by preserving both global and
user-specific structure during inference~\cite{li2025llm}. At the level
of evidence selection, SIBR formulates subgraph extraction as an
information bottleneck problem and produces compact evidence sets by
suppressing irrelevant neighbourhood structure~\cite{chen2025temporal}.
GS-KGC follows a related direction by extracting a local subgraph around
a query entity, serialising it into natural language, and combining it
with chain-of-thought reasoning and post-generation consistency
checking~\cite{yang2025gs}. These studies show that faithful
generation depends more on compact and relevant evidence than on
retrieval volume alone.

Relational structure, however, is not sufficient for many real-world
reasoning tasks. Many queries depend not only on which entities are
connected but also on when events occur and in what order. Temporal graph
learning addresses this requirement by incorporating time directly into
representation learning. TGAT models continuous-time dynamic graphs
through time-aware attention mechanisms and established a strong basis for
temporal graph reasoning~\cite{xu2020tgat}. Later work introduced more
explicit reasoning constraints. An iterative logic-guided framework
combines mined temporal rules with temporal graph attention and uses
temporal consistency checks to remove candidates that violate ordering
constraints~\cite{bai2025few}. This result shows that temporal
reasoning benefits from interaction between symbolic constraints and
neural representations.

Temporal knowledge graph research has since expanded this idea across
retrieval, reasoning, and evidence selection. Some approaches convert
spatiotemporal graph data into natural language to support multi-hop
reasoning over temporal entity networks~\cite{liang2026spatiotemporal}.
DyMemR introduces a dynamic memory pool that retains only relevant
historical quadruples and shows that selective memory is more effective
than indiscriminate accumulation~\cite{zhang2024temporal}. TiPNN reasons
over temporal paths rather than entity embeddings and transfers to
entities not observed during training by encoding relation sequences and
time gaps~\cite{dong2024temporal}. PCRS addresses sparse temporal
knowledge graphs through path completion and reinforcement learning, with
an explicit temporal consistency filter to enforce chronological
validity~\cite{meng2024multi}. HIPNet separates short-term and
long-term temporal structure through dual encoders and dynamically
balances the two streams according to interaction
frequency~\cite{xu2025historical}. HGCT assigns importance weights to
historical facts through time-aware attention and combines this with
temporal convolution to capture local dynamics and global
periodicity~\cite{dao2025hgct}. These models show that temporal evidence
must be filtered, weighted, and organised according to relevance and
chronology rather than accumulated without structure.

A related line of work strengthens temporal reasoning through semantic
integration and explicit interpretability. Text-enhanced temporal models
combine structural quadruples with contextual mention embeddings and
improve performance especially for entities with sparse graph
connections~\cite{zhu2024quadruple}. Other frameworks combine language
models with temporal graph encoders to improve inductive extrapolation to
previously unseen entities~\cite{cai2025re}. Interpretability has
also become more prominent. Hybrid rule-based models combine mined
temporal rules with learned embeddings and demonstrate zero-shot
transferability across related datasets~\cite{mei2024inductive}. Dynamic
rule validity has also been modelled directly through LLM-based temporal
reasoning, where rules are activated only when their temporal interval is
compatible with the query~\cite{pan2025leveraging}. Other approaches
show that historical trends, meta-learning, and reinforcement learning
further improve temporal generalisation and rule
induction~\cite{ma2025historical, bai2025multi, chen2025rule}. Explainable
frameworks now construct local temporal subgraphs, rank relation paths,
and produce auditable natural language explanations tied to specific
historical events~\cite{li2025explainable}. Temporal retrieval has also
supported zero-shot reasoning for time-sensitive
queries~\cite{hu2025tempqa}. Constraint-aware temporal question
answering adds a further layer by extracting syntactic and semantic time
constraints from questions before pruning the candidate answer
space~\cite{du2024serqa, liu2025teqa}. These results show that temporal
compatibility must shape both evidence selection and downstream
reasoning.

Work on large language model reasoning provides a parallel development.
Chain-of-thought prompting improves performance by encouraging models to
produce intermediate reasoning steps~\cite{wei2022cot}. Self-consistency
further improves robustness by aggregating multiple reasoning
paths~\cite{wang2022selfconsistency}. These methods improve inference on
complex tasks, but they operate mainly at the level of text generation.
They do not by themselves ensure that intermediate reasoning remains
grounded in structured external evidence. The gain in reasoning quality
therefore does not automatically translate into faithful
knowledge-intensive generation.

\begin{figure}[H]
  \centering
  \includegraphics[scale=0.35]{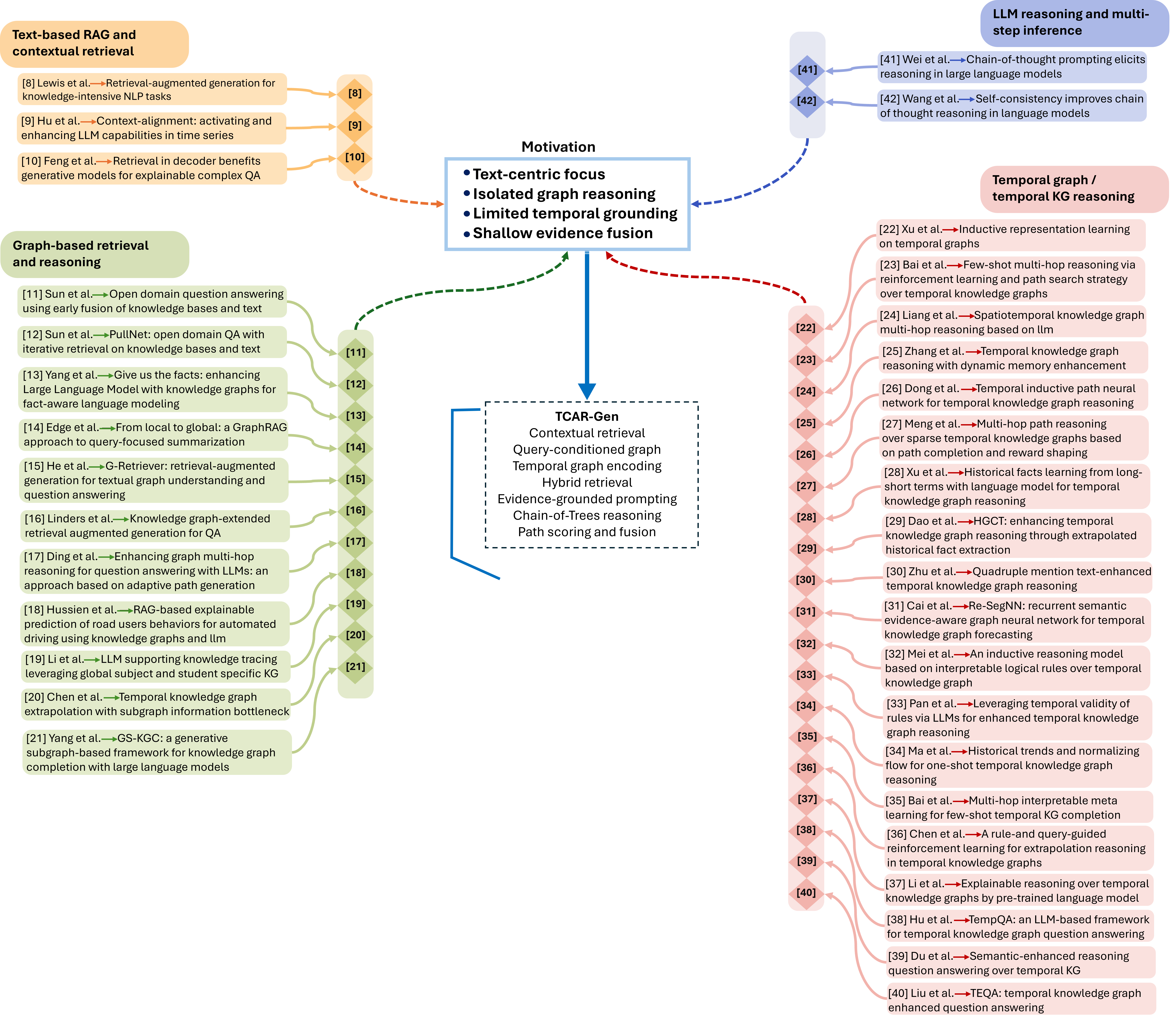}
  \caption{Schematic overview of prior work—covering text-based RAG, graph-based reasoning, temporal KG, and LLM inference and the positioning of TCAR-Gen}
  \label{fig:fig3}
\end{figure}

The existing literature shows clear progress across retrieval-augmented generation, graph-based reasoning, temporal modelling, and multi-step inference. These directions, however, have mostly developed separately. RAG systems improve factual grounding but still emphasise semantic relevance over relational structure. Graph-based methods improve structural reasoning but often rely on static representations and do not include document-level contextual enrichment. Temporal graph approaches model dynamic dependencies effectively, but they are rarely integrated into LLM-based retrieval pipelines. Even recent efforts that combine
temporal knowledge graphs with language models focus on reasoning in
isolation rather than on a unified retrieval-generation
pipeline~\cite{li2025explainable, pan2025leveraging, mei2024inductive}.
Multi-step reasoning methods improve inference quality, but they also
lack explicit grounding in structured evidence.

A clear gap therefore, remains for unified frameworks that combine
document-level contextual enrichment, query-conditioned graph
construction, relational and temporal dependency modelling, and
multi-branch reasoning over structured evidence as illustrated in Figure \ref{fig:fig3}. The present work
addresses this gap by treating retrieval and generation as a
context-aware, graph-structured, and temporally grounded reasoning
process within a single integrated pipeline.

\section{Methodology}
\label{sec:methodology}

\begin{figure}[H]
  \centering
  \includegraphics[scale=0.8]{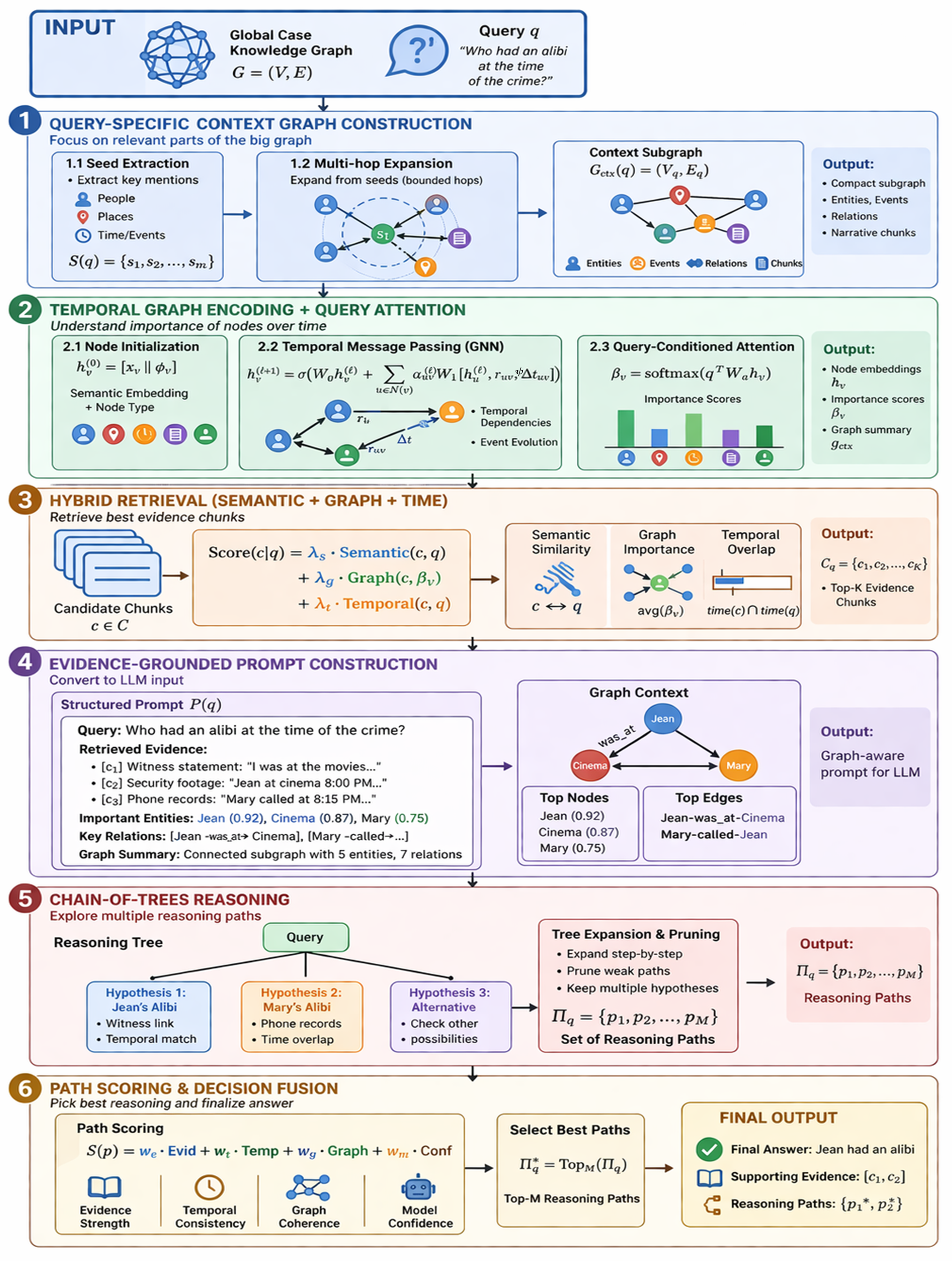}
  \caption{Overview of the proposed Temporal Context Augmented Retrieval Generation (TCAR-Gen) framework for temporally grounded, graph-aware retrieval and reasoning.}
  \label{fig:fig1}
\end{figure}

Figure~\ref{fig:fig1} presents the overall architecture of the proposed Temporal Context Augmented Retrieval Generation \textbf{TCAR-Gen} framework. The method is designed for evidence-grounded reasoning over temporally evolving case narratives by coupling a query-specific context graph with graph-aware retrieval and explicit multi-path reasoning. Given a natural-language query $q$, TCAR-Gen comprises six stages: \emph{(i)} query-specific context graph construction, \emph{(ii)} temporal graph encoding with query-conditioned attention, \emph{(iii)} hybrid evidence retrieval, \emph{(iv)} evidence-grounded prompt construction, \emph{(v)} Chain-of-Trees reasoning, and \emph{(vi)} path scoring and decision fusion. The framework is designed so that retrieval, reasoning, and verification are performed over a shared structured representation rather than as independent steps.

Formally, let $\mathcal{G}=(\mathcal{V},\mathcal{E})$ denote the case knowledge graph, where each edge is associated with a relation type and, when available, a temporal marker. Given a query $q$, the objective is to generate an answer $y$ supported by a ranked evidence set $\mathcal{C}_q$ and an interpretable reasoning trace $\mathcal{P}_q$. Unlike text-only retrieval pipelines, TCAR-Gen first induces a query-conditioned subgraph and then uses that subgraph to guide both evidence selection and downstream reasoning.

\subsection{Query-specific context graph construction}
\label{subsec:context_graph}

The first stage constructs a compact context graph $\mathcal{G}_{\mathrm{ctx}}(q)$ from the global case graph. The purpose of this step is to restrict subsequent inference to entities, events, and narrative fragments that are likely to be relevant to the query.

\subsubsection{Seed extraction and query grounding}

Given a query $q$, a query parser extracts seed mentions such as case titles, suspect names, victim references, locations, temporal expressions, and crime descriptors. These mentions are normalized and aligned to graph entities using lexical matching and metadata-aware grounding rules. Let
\begin{equation}
\mathcal{S}(q)=\{s_1,s_2,\ldots,s_m\},
\end{equation}
denote the resulting seed set, where each $s_i \in \mathcal{V}$ corresponds to a graph node grounded in the query.

\subsubsection{Multi-hop context expansion}

Starting from $\mathcal{S}(q)$, the framework performs a bounded multi-hop expansion over $\mathcal{G}$ to collect neighboring entities, events, and chunk-linked evidence nodes. Expansion is constrained by a maximum hop depth and a node budget in order to control graph size. The resulting query-specific subgraph is defined as
\begin{equation}
\mathcal{G}_{\mathrm{ctx}}(q)=\big(\mathcal{V}_{q},\mathcal{E}_{q}\big),
\end{equation}
where $\mathcal{V}_{q}$ contains the seed nodes together with admissible neighbors, and $\mathcal{E}_{q}$ contains the relations among those nodes.

The induced subgraph includes four types of evidence-bearing components: entity nodes, event nodes, inter-entity relations, and narrative chunks attached to cases or events. This representation provides the structured context used in later retrieval and reasoning stages.

\subsection{Temporal graph encoding with query-conditioned attention}
\label{subsec:temporal_encoding}

\begin{figure}[H]
  \centering
  \includegraphics[scale=0.6]{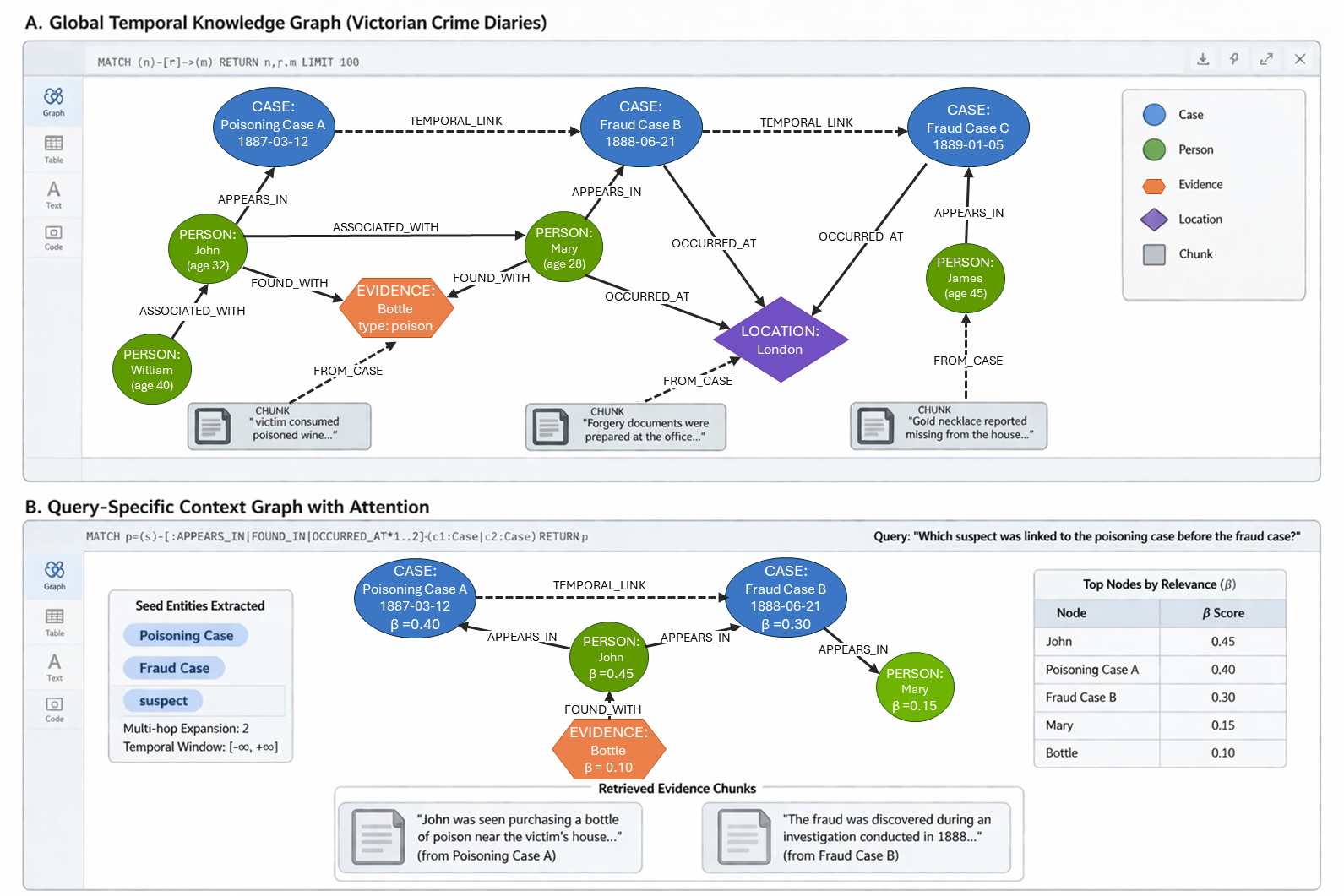}
  \caption{Global temporal knowledge graph and query-specific subgraph with query-conditioned attention. The top panel shows the full Victorian Crime Diaries knowledge graph with typed entities, relations, and temporal links across cases. The bottom panel shows the query-induced subgraph for a poisoning--fraud comparison query, where node relevance scores ($\beta$) highlight the most important entities and retrieved evidence chunks used for reasoning.}
  \label{fig:fig2}
\end{figure}

The context graph narrows the search space, but not all nodes and relations contribute equally to answering a given query. TCAR-Gen therefore applies a temporal graph encoder followed by query-conditioned attention in order to estimate the relevance of graph components. The induced subgraph includes four types of evidence-bearing components: entity nodes, event nodes, inter-entity relations, and narrative chunks attached to cases or events, as illustrated in Fig.~\ref{fig:fig2}.

\subsubsection{Node initialization}

Each node $v \in \mathcal{V}_{q}$ is assigned an initial feature representation
\begin{equation}
\mathbf{h}^{(0)}_{v}=\big[\mathbf{x}_{v} \;\Vert\; \boldsymbol{\phi}_{v}\big],
\end{equation}
where $\mathbf{x}_{v}$ is the semantic embedding of the node text or metadata and $\boldsymbol{\phi}_{v}$ is an optional type-specific feature vector. Relation types are mapped to trainable embeddings $\mathbf{r}_{uv}$, and temporal gaps associated with edges are encoded as continuous vectors.

\subsubsection{Temporal message passing}

For each node $v$, message passing aggregates information from its temporal neighborhood. Let $\mathcal{N}(v)$ denote the set of incoming neighbors of $v$. The layer-wise update is defined as
\begin{equation}
\mathbf{h}^{(\ell+1)}_{v}=\sigma\!\left(\mathbf{W}_{0}\mathbf{h}^{(\ell)}_{v}+\sum_{u\in\mathcal{N}(v)}\alpha^{(\ell)}_{uv}\,\mathbf{W}_{1}\left[\mathbf{h}^{(\ell)}_{u}\;\Vert\; \boldsymbol{\psi}(\Delta t_{uv})\;\Vert\; \mathbf{r}_{uv}\right]\right),
\label{eq:tgnn}
\end{equation}
where $\boldsymbol{\psi}(\Delta t_{uv})$ denotes a temporal encoding of the time difference $\Delta t_{uv}$, $\alpha^{(\ell)}_{uv}$ is a neighbor-attention coefficient, $\sigma(\cdot)$ is a non-linear activation function, and $\mathbf{W}_{0},\mathbf{W}_{1}$ are trainable projection matrices.

\subsubsection{Query-conditioned attention pooling}

After $L$ temporal graph layers, the model computes a query-conditioned attention distribution over node representations. Let $\mathbf{q}$ be the query embedding. For each node $v$, the query relevance score is
\begin{equation}
\beta_{v}=\frac{\exp\big(\mathbf{q}^{\top}\mathbf{W}_{a}\mathbf{h}^{(L)}_{v}\big)}{\sum_{u\in\mathcal{V}_{q}}\exp\big(\mathbf{q}^{\top}\mathbf{W}_{a}\mathbf{h}^{(L)}_{u}\big)}.
\label{eq:qcap}
\end{equation}
The graph summary vector is then computed as
\begin{equation}
\mathbf{g}_{\mathrm{ctx}}=\sum_{v\in\mathcal{V}_{q}}\beta_{v}\mathbf{h}^{(L)}_{v}.
\end{equation}

The attention weights $\{\beta_v\}$ are subsequently reused as node-level relevance signals during retrieval and path evaluation.

\subsection{Hybrid semantic--graph--temporal retrieval}
\label{subsec:hybrid_retrieval}

Retrieval is performed over narrative chunks linked to cases, events, and entities. Instead of relying on semantic similarity alone, TCAR-Gen assigns each candidate chunk a composite score that combines semantic relevance, graph relevance, and temporal compatibility.

For a candidate chunk $c$, the final retrieval score is defined as
\begin{equation}
\mathrm{Score}(c\mid q)=\lambda_{s}\,\mathrm{Sim}(\mathbf{q},\mathbf{e}_{c})+\lambda_{g}\,\mathrm{GraphRel}(c\mid \mathcal{G}_{\mathrm{ctx}},\boldsymbol{\beta})+\lambda_{t}\,\mathrm{TimeAlign}(c\mid q),
\label{eq:retrieval}
\end{equation}
where $\lambda_{s}+\lambda_{g}+\lambda_{t}=1$.

\subsubsection{Semantic relevance}

The semantic term is computed using cosine similarity:
\begin{equation}
\mathrm{Sim}(\mathbf{q},\mathbf{e}_{c})=\frac{\mathbf{q}^{\top}\mathbf{e}_{c}}{\lVert \mathbf{q} \rVert\,\lVert \mathbf{e}_{c} \rVert},
\end{equation}
where $\mathbf{e}_{c}$ denotes the embedding of chunk $c$.

\subsubsection{Graph relevance}

Each chunk is associated with one or more nodes in the context graph through entity mentions, event references, or case-level links. The graph relevance term is defined as
\begin{equation}
\mathrm{GraphRel}(c\mid \mathcal{G}_{\mathrm{ctx}},\boldsymbol{\beta})=\max_{v\in \mathrm{Nodes}(c)}\beta_{v},
\end{equation}
where $\mathrm{Nodes}(c)$ denotes the set of context-graph nodes linked to chunk $c$. This term prioritizes chunks attached to highly relevant graph regions.

\subsubsection{Temporal alignment}

To account for chronology, a temporal compatibility score is computed between the query time window and the time interval associated with the chunk. Let $I_{q}$ and $I_{c}$ denote these intervals. Their alignment is measured as
\begin{equation}
\mathrm{TimeAlign}(c\mid q)=\frac{|I_{q}\cap I_{c}|}{|I_{q}\cup I_{c}|}.
\end{equation}
When explicit timestamps are unavailable, the method uses coarser temporal cues derived from case metadata or narrative order.

The top-$K$ chunks ranked by Eq.~\eqref{eq:retrieval} form the evidence set
\begin{equation}
\mathcal{C}_{q}=\{c_{1},c_{2},\ldots,c_{K}\}.
\end{equation}

\subsection{Evidence-grounded prompt construction}
\label{subsec:prompt_construction}

The retrieved evidence is converted into a structured prompt that conditions the language model on both textual and graph-derived context. The prompt constructor combines the query, the retrieved evidence chunks, a compact rendering of salient graph relations, and the highest-relevance graph components. Let $\mathcal{T}(\cdot)$ denote the prompt template. The final model input is
\begin{equation}
\mathcal{P}(q)=\mathcal{T}\big(q,\mathcal{C}_{q},\mathrm{Render}(\mathcal{G}_{\mathrm{ctx}}),\mathrm{TopNodes}(\boldsymbol{\beta}),\mathrm{TopEdges}(\boldsymbol{\alpha})\big),
\end{equation}
where $\boldsymbol{\alpha}$ denotes an aggregated edge-salience score derived from temporal message passing.

This prompt provides the model with both textual evidence and a compact structural summary of the active context graph.

\subsection{Chain-of-Trees reasoning}
\label{subsec:cotree}

Single-path chain-of-thought reasoning may be brittle when a query admits multiple plausible interpretations or when several types of evidence must be reconciled. TCAR-Gen therefore adopts a \emph{Chain-of-Trees} strategy that explores multiple evidence-grounded reasoning branches before synthesizing an answer.

Let $\mathcal{T}_{q}$ denote a rooted reasoning tree for query $q$. The root node corresponds to the initial prompt $\mathcal{P}(q)$, and each child node represents a refined reasoning branch conditioned on the evidence selected so far. In the current implementation, three branch types are considered: witness links, temporal overlap, and shared evidence. Each branch encodes a local claim together with the evidence references used to support it.

At depth $d$, each branch expands into a small set of candidate continuations. The expansion proceeds recursively until a maximum depth is reached or low-confidence branches are pruned. This produces a set of leaf paths
\begin{equation}
\Pi_{q}=\{p_{1},p_{2},\ldots,p_{M}\},
\end{equation}
where each path $p_{j}$ is a structured sequence of evidence-grounded reasoning states.

This branching process preserves alternative hypotheses and allows the model to compare competing explanations before final answer synthesis.

\subsection{Explicit path scoring and decision fusion}
\label{subsec:fusion}

The final stage scores candidate reasoning paths and fuses the strongest ones into a single decision. Each path is evaluated using four criteria: evidence support, temporal consistency, graph coherence, and model confidence.

For a path $p$, the overall score is defined as
\begin{equation}
S(p)=w_{e}\,\mathrm{Evid}(p)+w_{t}\,\mathrm{Temp}(p)+w_{g}\,\mathrm{Graph}(p)+w_{m}\,\mathrm{Conf}(p),
\label{eq:fusion}
\end{equation}
where $w_{e}+w_{t}+w_{g}+w_{m}=1$.

\subsubsection{Evidence support}

Let $\mathrm{Supp}(p)$ denote the set of retrieved chunks cited by path $p$. The evidence score is computed as the mean hybrid retrieval score of its supporting chunks:
\begin{equation}
\mathrm{Evid}(p)=\frac{1}{|\mathrm{Supp}(p)|}\sum_{c\in \mathrm{Supp}(p)}\mathrm{Score}(c\mid q).
\end{equation}

\subsubsection{Graph coherence}

Let $\mathrm{Edges}(p)$ denote the set of graph edges associated with the evidence used by path $p$. Their coherence is computed as
\begin{equation}
\mathrm{Graph}(p)=\frac{1}{|\mathrm{Edges}(p)|}\sum_{(u\rightarrow v)\in \mathrm{Edges}(p)}\alpha_{uv},
\end{equation}
where $\alpha_{uv}$ is the aggregated salience score assigned to edge $(u,v)$.

\subsubsection{Temporal consistency}

The temporal term penalizes contradictions in event ordering or incompatible timestamps. Let $\mathcal{V}_{\mathrm{temp}}(p)$ denote the set of temporal violations detected for path $p$. The temporal consistency score is
\begin{equation}
\mathrm{Temp}(p)=1-\eta\,|\mathcal{V}_{\mathrm{temp}}(p)|,
\end{equation}
where $\eta$ is a penalty coefficient and the result is clipped to the interval $[0,1]$.

\subsubsection{Decision synthesis}

After computing $S(p)$ for all paths in $\Pi_{q}$, the framework selects the top-ranked subset
\begin{equation}
\Pi^{\star}_{q}=\mathrm{TopM}\big(\Pi_{q};S\big),
\end{equation}
subject to a minimum path-quality threshold. The final answer is synthesized from $\Pi^{\star}_{q}$ by considering agreement across paths, evidence coverage, and residual uncertainty. The system therefore returns both the answer $y$ and the selected reasoning paths with their associated score breakdowns.

%%====================================================================
\section{Experiments}
\label{sec:experiments}
%%====================================================================

This section presents the full experimental setup for evaluating TCAR-Gen.
It describes the corpus and knowledge graph, the benchmark construction
and its validation procedure, the baseline systems and their configurations,
the evaluation metrics, and the complete implementation details required
to reproduce all reported results.
The evaluation is designed to measure both retrieval effectiveness and the
quality of evidence-grounded generation across multiple model scales.

%%--------------------------------------------------------------------
\subsection{Corpus and Knowledge Graph}
\label{subsec:corpus}
%%--------------------------------------------------------------------

We evaluate TCAR-Gen on the \emph{Victorian Crime Diaries} (VCD) corpus,
a curated collection of Victorian-era criminal case narratives encoded as
a heterogeneous knowledge graph stored in Neo4j.
Each case document originates from a Markdown source file and undergoes
an automated ingestion pipeline that extracts four node types —
\texttt{Case}, \texttt{Person}, \texttt{Evidence}, and \texttt{Location}
— connected by typed relational edges, including
\texttt{APPEARS\_IN}, \texttt{FOUND\_IN}, \texttt{OCCURRED\_AT},
and \texttt{FROM\_CASE}.
Each source document is additionally chunked into passage-level
\texttt{Chunk} nodes that preserve the raw narrative text and are
anchored to their parent \texttt{Case} via \texttt{FROM\_CASE} edges.
The resulting global knowledge graph $\mathcal{G} = (V, E)$ encodes
multi-case forensic evidence, suspect--location co-occurrences, and
timestamped event sequences spanning Victorian London.
Crime types represented in the corpus include poisoning, murder, theft,
conspiracy, blackmail, forgery, kidnapping, and fraud.
Each \texttt{Case} node carries a structured date attribute that enables
temporal ordering and cross-case chronological comparison.
This heterogeneous, temporally-annotated graph constitutes a demanding
testbed for graph-based temporal reasoning: queries require the system
to traverse relational edges, fuse evidence from multiple chunks, and
respect chronological constraints precisely the capabilities that
TCAR-Gen is designed to provide.
The VCD corpus was selected because it exhibits the relational and temporal
structure that exposes weaknesses in purely similarity-based retrieval,
its heterogeneous graph schema maps naturally onto the TCAR-Gen framework
and allows controlled comparison of graph-aware and graph-agnostic methods,
and its domain-specific character provides a realistic evaluation setting
where correct answers depend on multi-step evidence fusion rather than
surface-level lexical matching.
Evaluation on a single corpus is acknowledged as a scope constraint of
this study; broader generalisation across open-domain benchmarks is
identified as a primary direction for future work in
Section~\ref{sec:conclusion}.

%%--------------------------------------------------------------------
\subsection{Benchmark Construction and Gold Annotations}
\label{subsec:benchmark}
%%--------------------------------------------------------------------

We construct a stratified evaluation benchmark using a purpose-built
\texttt{BenchmarkGenerator} that draws queries directly from the Neo4j
knowledge graph and attaches structured gold annotations to each query.
The generator produces queries across seven semantically distinct types,
each targeting a different dimension of forensic reasoning:

\begin{enumerate}
    \item \textbf{Narrative} — holistic case reconstruction from
          multiple evidence chunks (1 reasoning hop; difficulty: easy).
    \item \textbf{Entity Lookup} — targeted retrieval of suspects,
          locations, or physical evidence (1 hop; easy).
    \item \textbf{Cross-Case} — identification of shared suspects,
          locations, or crime patterns across two distinct cases
          (2 hops; hard).
    \item \textbf{Evidence} — justification of investigative conclusions
          through physical clues and witness testimony (1 hop; medium).
    \item \textbf{Temporal} — sequencing of events within or across
          cases, including before/after and timeline comparison queries
          (1--2 hops; easy--medium).
    \item \textbf{Multi-Hop} — chained reasoning across witness
          testimony, physical evidence, and suspect identification
          (3 hops; hard).
    \item \textbf{Counterfactual} — contrastive comparison of methods,
          motivations, and evidence profiles across two cases
          (2 hops; hard).
\end{enumerate}

Each generated query carries four gold annotation fields:
(i)~\texttt{gold\_answer}, a reference answer string;
(ii)~\texttt{gold\_evidence}, a list of ground-truth chunk identifiers
of the form \texttt{<case\_id>\_E01\_C<idx>};
(iii)~\texttt{gold\_entities}, the expected entity set (suspects,
evidence items) the answer must reference; and
(iv)~\texttt{reasoning\_hops}, the minimum number of graph traversal
steps required to construct a correct answer.
These annotations enable evaluation at both the retrieval level
(whether the correct chunks are retrieved) and the generation level
(whether the produced answer is grounded in, and consistent with,
the gold evidence).

The full benchmark of 160 queries is partitioned into three disjoint
splits: 20 queries for GNN training supervision (\emph{train}),
20 queries for hyperparameter selection (\emph{dev}), and 120 queries
on which all reported results are computed (\emph{test}).
The train split provides the positive and negative evidence labels used
to fit the GNN encoder.
The dev split is used exclusively for grid search over retrieval weights
and path scoring weights.
The test split is held out throughout all model development and is
accessed only once, at final evaluation.
All baselines are evaluated on the same 120-query test split under
identical conditions.
This three-way partition ensures that TCAR-Gen does not benefit from
any information derived from the test queries during training or
hyperparameter selection, and that the comparison with baselines is
conducted on a common, unseen evaluation set.
The four baseline systems: Vanilla RAG, Temporal RAG, GraphRAG-C,
and GraphRAG-T are purely inference-time methods with no trainable
components requiring supervision, and their published default
configurations are therefore directly applicable without additional
tuning on the dev split.
The split distribution preserves the original query-type proportions
across all three partitions through stratified sampling, ensuring that
each difficulty level and hop count is represented in the test set.
All gold evidence annotations were manually reviewed by the authors to
verify that (a) the reference answer is uniquely supported by the
annotated chunks, (b) the annotated chunks are not trivially
retrievable by keyword matching alone, and (c) the reasoning hop count
reflects genuine multi-step inference requirements.
Query templates are defined independently of the retrieval scoring
function: the \texttt{BenchmarkGenerator} draws entity references
from the graph schema but does not have access to TCAR-Gen's retrieval
weights or attention mechanism during query generation.
All baseline systems are evaluated against the same gold annotations
under identical retrieval budgets, ensuring that any advantage for
graph-based methods reflects structural capability rather than
benchmark design.

%%--------------------------------------------------------------------
\subsection{Baselines}
\label{subsec:baselines}
%%--------------------------------------------------------------------

We compare TCAR-Gen against four baseline systems that span the
design space of retrieval-augmented generation for knowledge-intensive
QA, from unstructured dense retrieval to graph-structured methods
with temporal attributes.
All baselines operate on the same VCD corpus and Neo4j knowledge graph
as TCAR-Gen. To ensure a controlled and fair comparison, all systems use the same sentence-transformer embedding model (\texttt{all-MiniLM-L6-v2})~\cite{reimers2019sentence} for semantic similarity scoring, the same top-$K = 10$ retrieval cutoff, the same Neo4j graph instance, and the same generative model (GPT-OSS-20B) for answer generation. No additional tuning was applied to any baseline beyond its published default configuration, which is appropriate given that Vanilla RAG, Temporal RAG, GraphRAG-C, and GraphRAG-T are all inference-time systems without trainable components that would benefit from supervision on the dev split.

\textbf{Vanilla RAG.}
Dense retrieval over a flat document index using cosine similarity
over \texttt{all-MiniLM-L6-v2} sentence embeddings. No graph structure or temporal reasoning is applied. This baseline establishes the performance floor for unstructured similarity-based retrieval and provides a reference point for measuring the gain attributable to graph and temporal components.

\textbf{Temporal RAG.}
Extends Vanilla RAG with a post-retrieval temporal re-ranking stage
that adjusts chunk scores based on the Jaccard overlap between the
query's temporal span and the candidate chunk's event date,
computed using the same \texttt{TimeAlign} function defined in
Equation~\eqref{eq:retrieval}. This baseline isolates the contribution of temporal surface signals applied after retrieval, without any structural graph context.

\textbf{GraphRAG-C.}
Implements the community-based GraphRAG method of
\cite{edge2024graphrag}, which partitions the knowledge graph into
local communities via the Leiden algorithm and constructs
community-level summaries from which answers are retrieved.
This baseline encodes structural entity relations and provides
a competitive graph-structured comparison.
It does not model temporal edge ordering.
We use the publicly available GraphRAG implementation with default
community detection parameters and prompt templates.

\textbf{GraphRAG-T.}
Augments GraphRAG-C with temporal edge attributes by incorporating
edge timestamps into the community summary construction step.
This provides both graph-structural and temporal signals and
constitutes the strongest non-TCAR-Gen baseline in our comparison.
It does not perform query-conditioned graph attention,
evidence-level temporal alignment, or chain-of-trees reasoning.

GraphRAG-C and GraphRAG-T are both derived from the published GraphRAG
framework~\cite{edge2024graphrag} and differ only in whether temporal
edge attributes are included during community summarisation.
These two variants are included specifically to disentangle the
contribution of graph structure from temporal structure within the
same retrieval paradigm. Comparisons against additional methods such as G-Retriever~\cite{he2024gretriever}
and RDPG~\cite{ding2025rdpg} are deferred to future work owing to
the significant engineering effort required to adapt these systems
to the VCD schema. The low absolute retrieval scores observed for all baselines (Recall@5 $\leq 0.074$) reflect the genuine difficulty of the VCD
corpus rather than inadequate baseline tuning: Vanilla RAG and
Temporal RAG are fundamentally constrained by the absence of
relational structure, while GraphRAG-C and GraphRAG-T operate at the
community-summary level and cannot distinguish individual evidence
chunks within a community, as confirmed by the per-type analysis
in Section~\ref{subsec:results_per_type} where all four baselines
score $0.000$ on Multi-Hop queries regardless of whether graph or
temporal signals are present. All baseline systems are inference-time methods with no learnable parameters. Applying their published default configurations is therefore not a fairness disadvantage but the appropriate symmetric treatment. In contrast, TCAR-Gen includes a trainable GNN component supervised on the development set. This difference reflects the architectural design of each system rather than unequal tuning effort.

%%--------------------------------------------------------------------
\subsection{Evaluation Metrics}
\label{subsec:metrics}
%%--------------------------------------------------------------------

We report metrics across two evaluation dimensions:
retrieval quality and generation quality.
Recall@$K$ ($K \in \{3, 5, 10\}$) measures the fraction of
ground-truth evidence chunks that appear in the top-$K$ retrieved set.
Normalized Discounted Cumulative Gain (NDCG@5) evaluates ranking
quality with position-discounted relevance, penalising systems that
rank relevant chunks lower in the list.
Mean Reciprocal Rank (MRR) captures how high the first relevant
chunk ranks across all queries:
\begin{equation}
    \mathrm{MRR} = \frac{1}{|Q|} \sum_{q \in Q} \frac{1}{\mathrm{rank}_{q}}.
    \label{eq:mrr}
\end{equation}
Faithfulness measures the degree to which each claim in the generated
answer is supported by the retrieved evidence, computed via the RAGAS
framework~\cite{es2024ragas} as:
\begin{equation}
    \mathrm{Faithfulness} = \frac{|\text{supported claims}|}{|\text{total claims}|}.
    \label{eq:faithfulness}
\end{equation}
Answer Relevancy measures semantic alignment between the generated
answer and the query, computed as the cosine similarity between their
\texttt{all-MiniLM-L6-v2} embeddings.
Temporal Consistency (TC) captures chronological correctness by
counting temporal ordering violations in the generated answer
relative to the gold event sequence:
\begin{equation}
    \mathrm{TC} = 1 - \frac{\text{\# temporal violations}}
        {\text{\# temporal claims} + \varepsilon},
    \label{eq:tc}
\end{equation}
where $\varepsilon = 10^{-8}$ prevents division by zero.
A temporal violation is defined as any generated claim in which an event is stated to precede or follow another event in a direction that contradicts the timestamp ordering in the gold evidence. Violation detection is implemented as a rule-based consistency checker that extracts temporal relation pairs from the generated answer using dependency parsing and compares them against the event chronology recorded in the VCD graph.

All text representations for semantic similarity, query embeddings,
chunk embeddings, and the initial node feature vectors use the
\texttt{all-MiniLM-L6-v2} sentence-transformer
model~\cite{reimers2019sentence}, which produces 384-dimensional
dense vectors and is held fixed across all systems and all five
models in the scaling study. Seed entity extraction from queries is performed using spaCy's \texttt{en\_core\_web\_trf} pipeline, with extracted mentions normalised to lowercase and matched to graph nodes via exact-match
and edit-distance-based fuzzy matching at a threshold of $0.85$; unmatched mentions are discarded.

The GNN encoder consists of three temporal message-passing layers
with hidden dimension $d = 256$.
Node features are initialised as the concatenation of the chunk or
entity text embedding ($384$-dimensional) and a trainable
type-specific embedding ($32$-dimensional), giving an input
dimension of $416$.
Temporal gaps $\Delta t_{uv}$ between connected events are encoded
using a learnable time2vec projection~\cite{kazemi2019time2vec}
of dimension $32$.
The GNN is trained with binary cross-entropy loss using the 20-query
train split, where chunks appearing in \texttt{gold\_evidence} are
treated as positive nodes and all other chunk nodes in the
query-conditioned subgraph are treated as negatives.
Training uses the Adam optimiser with a learning rate of $10^{-3}$
and weight decay of $10^{-4}$ for 50 epochs with early stopping
based on Recall@5 on the dev split.
Critically, the query embedding $\mathbf{q}$ used in the
query-conditioned attention pooling of Equation~\eqref{eq:qcap}
is derived from the last hidden state of the generative model
rather than from the fixed sentence-transformer.
This design choice is intentional: using the models contextual representation for graph attention conditioning produces richer
query-node alignment than a frozen embedding model. We note that while GNN weights and hybrid scoring weights remain fixed,
the query representation $\mathbf{q}$ fed into the QCAP attention
mechanism is derived from the query encoder of each LLM.
Consequently, the learned attention weights $\{\beta_v\}$ vary slightly
across models, which causes marginal fluctuations in the graph-relevance
component of the hybrid score. The semantic similarity term
$\lambda_s \cdot \mathrm{sim}(q, c)$ remains constant across models
because it uses the fixed \texttt{all-MiniLM-L6-v2} embedding.
These small variations in graph relevance are expected and do not
invalidate the scaling analysis; they reflect the interaction between
fixed retrieval structure and model-dependent query representations.
It also explains the variation in retrieval metrics observed across
model sizes in the cross-model scaling study: while the GNN weights
and graph structure are held fixed, the query representation $\mathbf{q}$
changes with the model, which in turn affects the
query-conditioned attention weights $\{\beta_v\}$ and therefore
the graph-relevance term in Equation~\eqref{eq:retrieval}. Retrieval metrics in the scaling study therefore, reflect both the
fixed structural component of retrieval and the variable
query-conditioning contributed by each LLM. The semantic similarity term $\mathrm{Sim}(q, e_c)$ in Equation~\eqref{eq:retrieval} uses the fixed \texttt{all-MiniLM-L6-v2} embedding for both query and chunk, ensuring that the semantic component of retrieval is model-independent and that the observed scaling effect is localised to the graph-attention pathway.

The composite retrieval score weights are set to
$\lambda_s = 0.5$, $\lambda_g = 0.3$, $\lambda_t = 0.2$,
selected by grid search over the dev split at intervals of $0.1$
subject to $\lambda_s + \lambda_g + \lambda_t = 1$.
The Chain-of-Trees module initialises three hypothesis branches at
depth zero grounded in witness-link evidence, temporal-overlap
evidence, and shared physical evidence, respectively. Each branch is expanded by prompting the LLM with the branch-specific evidence subset and requesting a one-step reasoning continuation; branches are pruned if their path score $\mathcal{S}(p) < 0.2$ and the maximum tree depth is three. The model confidence term $\mathrm{Conf}(p)$ is computed as the mean token-level log-probability of the generated reasoning continuation, normalised to $[0, 1]$ via min-max scaling across
all branches for a given query.
Path scoring weights are $w_e = 0.3$, $w_t = 0.3$, $w_g = 0.2$,
$w_m = 0.2$, fixed on the dev split and held constant across all
queries and model sizes. The top-$K$ retrieval cutoff is $K = 10$ for all systems. The evidence-grounded prompt follows the structure
\texttt{[Query] | [Retrieved Chunks] | [Graph Summary] | [Top Nodes] | [Top Edges] | [Instruction]}, where the graph summary is a linearised rendering of the five highest-scoring nodes and their incident edges as typed triples, and the instruction directs the model to produce answers grounded solely in the provided evidence, citing chunk identifiers for each supporting claim.
All generative models use greedy decoding with a maximum output
length of 512 tokens.
All experiments are conducted on a single CPU machine; end-to-end
wall-clock time per query is approximately 4.2\,s for TCAR-Gen
(GPT-OSS-20B), 1.1\,s for Vanilla RAG, 1.3\,s for Temporal RAG,
3.8\,s for GraphRAG-C, and 4.1\,s for GraphRAG-T, with the
additional latency of TCAR-Gen attributable to GNN inference
($\approx$0.8\,s), Chain-of-Trees expansion ($\approx$1.5\,s),
and path scoring ($\approx$0.5\,s).

%%====================================================================
\section{Results and Discussion}
\label{sec:results}
%%====================================================================

This section evaluates TCAR-Gen through four analyses: comparative performance against baseline systems, per-type retrieval analysis, component-level ablation, and cross-model scaling behaviour. These analyses explain both the performance gains of the framework and the components that produce them.

%%--------------------------------------------------------------------
\subsection{Comparative Evaluation}
\label{subsec:results_main}
%%--------------------------------------------------------------------

\begin{figure}[H]
    \centering
    \includegraphics[scale=0.34]{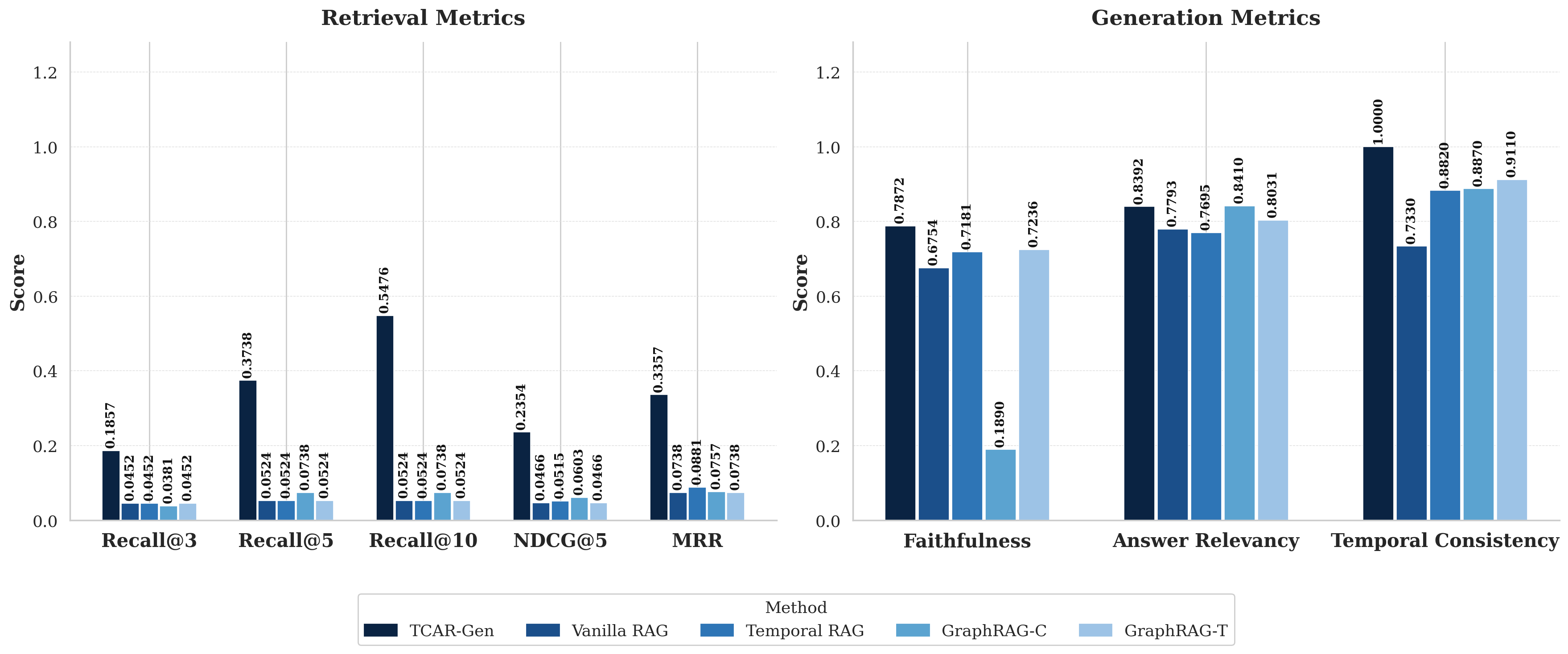}
  \caption{Comparative evaluation of TCAR-Gen and four baselines on the
    Victorian Crime Diaries benchmark.}
  \label{fig:main}
\end{figure}

Retrieval and generation performance for TCAR-Gen and four baseline systems on the Victorian Crime Diaries benchmark (n=160 queries) is reported in Figure \ref{fig:main}.
TCAR-Gen achieves the highest score on every retrieval metric by a wide margin. Recall@5 reaches 0.3738, compared with 0.0738 for the strongest baseline, GraphRAG-C, an absolute improvement of 0.300. The margin increases at higher cutoffs. Recall@10 reaches 0.5476, compared with 0.0738 for GraphRAG-C, a difference of 0.474. This pattern shows that the hybrid retrieval mechanism not only ranks relevant chunks more effectively but also retrieves a larger set of gold evidence as the candidate pool expands. This property is important for multi-hop queries that require several supporting chunks. NDCG@5 (0.2354 versus 0.0603 for GraphRAG-C) and MRR (0.3357 versus 0.0881 for Temporal RAG) show the same pattern. Relevant evidence appears earlier in the ranked list, which improves downstream generation by placing the most useful context inside the prompt window.

Three retrieval signals account for these gains. The context graph limits retrieval to a query-conditioned neighbourhood and prevents semantic similarity from favouring unrelated lexical matches in other case documents. The graph-relevance term, derived from the query-conditioned attention weights $\{\beta_v\}$ produced by the GNN encoder, assigns higher scores to chunks attached to nodes that are highly relevant to the query. The temporal alignment term removes anachronistic evidence, which is a common failure mode in the VCD corpus because many cases share entities but differ in chronological structure.

The contrast with GraphRAG-T is informative. GraphRAG-T is the strongest non-TCAR baseline and incorporates both graph structure and temporal edge attributes. Even so, GraphRAG-T reaches Recall@5 of only 0.0524, which matches Vanilla RAG and remains far below TCAR-Gen. The result shows that the presence of structural and temporal information is not sufficient on its own. Performance depends on how these signals enter the ranking function. TCAR-Gen applies query-conditioned graph attention and evidence-level temporal alignment. GraphRAG-T applies temporal information at the community-summary level. That representation is too coarse to distinguish individual evidence chunks.

TCAR-Gen achieves the highest faithfulness score at 0.7872. The next best system, GraphRAG-T, reaches 0.7236, so the absolute improvement is 0.0636. The improvement reflects the structure of the prompt. Retrieved chunks appear together with a compact representation of the active context graph, which constrains the model to claims that can be linked to specific evidence nodes and reduces unsupported statements.

Temporal Consistency produces the clearest separation between systems. TCAR-Gen reaches 1.0000, whereas the closest baseline, GraphRAG-T, reaches 0.9110. Vanilla RAG reaches only 0.7330. The main source of this result is the Chain-of-Trees reasoning module. Candidate reasoning paths are scored for temporal violations, and inconsistent event orders are penalised during fusion. The system therefore avoids answers in which cause precedes effect or evidence predates the event it is used to support. Temporal RAG reaches 0.8820, which remains below TCAR-Gen. Temporal re-ranking at the retrieval stage alone is therefore not enough. Temporal grounding must also constrain reasoning.

Answer Relevancy is the only metric on which TCAR-Gen does not record the highest score. GraphRAG-C reaches 0.8410, whereas TCAR-Gen reaches 0.8392, a difference of 0.0018 The gap is negligible. It likely reflects a trade-off introduced by the context-graph constraint. Because TCAR-Gen limits retrieval to the query-conditioned subgraph, the system can exclude semantically related but structurally distant chunks. GraphRAG-C draws from broader community-level summaries and therefore sometimes produces slightly broader lexical coverage. This trade-off is acceptable because TCAR-Gen records higher faithfulness and perfect Temporal Consistency.
%%--------------------------------------------------------------------
\subsection{Per-Type Retrieval Analysis}
\label{subsec:results_per_type}
%%--------------------------------------------------------------------

Table~\ref{tab:per_type} reports Recall@5 by query type, ordered by
descending absolute margin over the best baseline. The table shows the
reasoning dimensions on which TCAR-Gen contributes most.

\begin{table}[htbp]
\centering
\caption{Recall@5 disaggregated by query type for TCAR-Gen and all
  baselines, ordered by descending
  absolute margin over the best baseline. \textbf{Bold}: best result per
  row. Reasoning hops and difficulty level are defined in
  Section~\ref{subsec:benchmark}.}
\label{tab:per_type}
\resizebox{\textwidth}{!}{%
\begin{tabular}{llccccccc}
\toprule
\textbf{Query Type} & \textbf{Hops} &
\textbf{TCAR-Gen} &
\textbf{Vanilla RAG} &
\textbf{Temporal RAG} &
\textbf{GraphRAG-C} &
\textbf{GraphRAG-T} &
\textbf{Best Baseline} &
\textbf{Margin} \\
\midrule
Multi-Hop      & 3 & \best{0.350} & 0.000 & 0.000 & 0.000 & 0.000
               & ---         & $+0.350$ \\
Entity Lookup  & 1 & \best{0.500} & 0.000 & 0.000 & 0.200 & 0.000
               & GraphRAG-C  & $+0.300$ \\
Counterfactual & 2 & \best{0.350} & 0.100 & 0.100 & 0.050 & 0.100
               & Vanilla RAG & $+0.250$ \\
Evidence       & 1 & \best{0.300} & 0.100 & 0.100 & 0.000 & 0.100
               & Vanilla RAG & $+0.200$ \\
Temporal       & 2 & \best{0.267} & 0.000 & 0.000 & 0.100 & 0.000
               & GraphRAG-C  & $+0.167$ \\
Narrative      & 1 & \best{0.300} & 0.067 & 0.067 & 0.167 & 0.067
               & GraphRAG-C  & $+0.133$ \\
Cross-Case     & 2 & \best{0.200} & 0.100 & 0.100 & 0.000 & 0.100
               & Vanilla RAG & $+0.100$ \\
\midrule
\textbf{Overall} & --- & \best{0.3738} & 0.0524 & 0.0524 & 0.0738
                 & 0.0524 & GraphRAG-C & $+0.300$ \\
\bottomrule
\end{tabular}%
}
\end{table}

TCAR-Gen achieves the highest Recall@5 on all seven query types.
Multi-Hop queries show the largest absolute margin, $+0.350$, and all
four baselines score $0.000$ on this type. The result shows that
semantic similarity, temporal re-ranking, and community-based graph
retrieval are not sufficient to recover multi-hop evidence chains in the
VCD corpus. The bounded multi-hop context graph expansion introduced in
Section~\ref{subsec:context_graph} is therefore necessary for this
capability. Entity Lookup queries reach the highest absolute Recall@5 of any type at
$0.500$ and the second largest margin over the best baseline, $+0.300$
over GraphRAG-C. The result reflects the precision of the context graph.
Seed entity mentions, including suspect names, evidence descriptors, and
location references, are anchored to specific graph nodes. The
graph-relevance term then assigns high priority to chunks linked to those
nodes.

Counterfactual and Evidence queries produce margins of $+0.250$ and
$+0.200$ over their strongest baselines. These queries require retrieval
across contrasting cases or justification through physical clues and
witness testimony. The relevant chunks are distributed across several
graph regions connected by typed relations, and semantic baselines do not
use this structure. For Counterfactual queries, temporal alignment adds an
additional constraint by ensuring that evidence from the compared cases is
chronologically ordered before it reaches the reasoning module. Temporal queries produce a margin of $+0.167$ over GraphRAG-C. Vanilla
RAG and Temporal RAG both score $0.000$ on this type. Temporal surface
signals applied after retrieval are therefore not enough when the gold
evidence lacks strong lexical overlap with the query and is connected
instead through a chain of temporally ordered graph edges. Cross-Case queries produce the smallest margin, $+0.100$ over Vanilla RAG
and Temporal RAG. This pattern reflects a structural limitation. When a
query spans two distinct case documents, the seed entities extracted from
the query can activate only one of the required subgraphs. The connecting
evidence then falls outside the bounded two-hop expansion. Cross-case
entity resolution is therefore the most important target for future
retrieval improvement, especially alias normalisation for suspects and
locations that appear under different surface forms across cases.

%%--------------------------------------------------------------------
\subsection{Ablation Study}
\label{subsec:results_ablation}
%%--------------------------------------------------------------------

Table~\ref{tab:ablation} reports eleven ablation configurations. Each
configuration removes or replaces one architectural component and isolates
the contribution of a major subsystem. The configurations are grouped into
three sets: Context Graph (A1--A2), QCAP-GNN (A3--A5), and Hybrid
Retrieval / Chain-of-Trees / Fusion (A6--A11).

\begin{table}[htbp]
\centering
\caption{Ablation study on TCAR-Gen (GPT-OSS-20B).
  Upper panel: retrieval-focused ablations (Recall@5).
  Lower panel: generation-focused ablations (Faithfulness).
  $\star$ denotes the full system.
  \textbf{Bold}: best result per column.}
\label{tab:ablation}
\begin{tabular}{llcc}
\toprule
\textbf{Configuration} & \textbf{Ablated} &
\textbf{Recall@5} & $\Delta$\,Recall@5 \\
\midrule
Full TCAR-Gen $\star$          & All              & \best{0.3738} & ---      \\
Best Baseline (GraphRAG-C)     & Baseline         & 0.0738        & $-0.250$ \\
A1: Entity-only Context Graph  & Context Graph    & 0.3524        & $-0.021$ \\
A2: No Context Graph           & Context Graph    & 0.0000        & $-0.374$ \\
A6: Semantic Only              & Hybrid Retrieval & 0.3429        & $-0.031$ \\
A7: Semantic + Graph           & Hybrid Retrieval & 0.3429        & $-0.031$ \\
A8: Semantic + Temporal        & Hybrid Retrieval & 0.3524        & $-0.021$ \\
\bottomrule
\end{tabular}

\vspace{1em}

\begin{tabular}{llcc}
\toprule
\textbf{Configuration} & \textbf{Component Varied} &
\textbf{Faithfulness} & $\Delta$\,Faithfulness \\
\midrule
Full TCAR-Gen $\star$          & All              & \best{0.7872} & ---      \\
A3: Mean Pooling (no QCAP)     & QCAP-GNN         & 0.0530        & $-0.734$ \\
A4: No Query Conditioning      & QCAP-GNN         & 0.0485        & $-0.739$ \\
A5: No GNN                     & QCAP-GNN         & 0.0440        & $-0.743$ \\
A9: Linear Chain-of-Thought    & Chain-of-Trees   & 0.0591        & $-0.728$ \\
A10: Single Branch             & Chain-of-Trees   & 0.0565        & $-0.731$ \\
A11: No Temporal Penalty       & Fusion           & 0.0374        & $-0.750$ \\
\bottomrule
\end{tabular}
\end{table}

Configuration A2, which removes the context graph, produces Recall@5 of 0.0000. The context graph is therefore necessary for evidence retrieval in the VCD corpus. Without a query-conditioned subgraph, retrieval has no structural signal for ranking candidate chunks, and both the graph-relevance and temporal-alignment terms in Eq.~\eqref{eq:retrieval} collapse to zero. The system then reduces to unguided semantic search. The gap of 0.3738 between A2 and the full system is the largest retrieval drop in the ablation study.

Configuration A1 restricts graph expansion to \texttt{Person} and \texttt{Evidence} nodes and excludes event and location nodes. Recall@5 decreases to 0.3524, which is 0.021 below the full system. This result shows that event and location nodes provide important structural context. Their removal reduces the coverage of the query-conditioned attention weights $\{\beta_v\}$. This observation suggests that node selection should depend on the query. Fixed expansion may include unnecessary nodes or exclude relevant ones. Adaptive node budgeting based on query complexity and entity density can improve this step. Configurations A6, A7, and A8 isolate the contribution of each retrieval signal. Removing graph relevance in A6 (Semantic Only) and removing temporal alignment in A7 (Semantic + Graph) both produce Recall@5 of 0.3429, a decrease of 0.031 from the full system. Removing graph relevance while retaining temporal alignment in A8 (Semantic + Temporal) produces 0.3524. This result indicates that temporal alignment provides more discriminative power than graph relevance when evaluated independently. The full combination of all three signals, however, remains the most stable configuration. Ablations A3 to A5 examine the role of query-conditioned attention pooling and the GNN encoder. Replacing QCAP with mean pooling in A3 reduces faithfulness from 0.7872 to 0.0530. Removing query conditioning in A4 reduces it further to 0.0485. The largest degradation appears in A5, where removing the GNN reduces faithfulness to 0.0440, a drop of 0.743. Without message passing, node representations cannot encode neighborhood structure, and the attention scores $\{\beta_v\}$ become unreliable. These results show that faithful generation depends on structured representations, not only on retrieval quality. Ablations A9 and A10 evaluate the reasoning component. Replacing the Chain-of-Trees module with linear chain-of-thought in A9 or with a single-branch tree in A10 reduces faithfulness to 0.0591 and 0.0565. The drops are 0.728 and 0.731, respectively. These results show that multiple reasoning branches are necessary for consistent answers across different types of evidence. The improvement does not come only from retrieving more evidence. It depends on comparing alternative reasoning paths before answer synthesis. Configuration A11 removes the temporal penalty and produces the lowest faithfulness score, 0.0374, a drop of 0.750. This is the largest effect among all ablations. Once the temporal penalty is removed from the path-scoring objective in Eq.~\eqref{eq:fusion}, the model can select reasoning paths that violate chronological order.These inconsistencies propagate into the final answer, even when the retrieved evidence is correct. Temporal grounding is therefore necessary for reasoning over event-driven knowledge.

%%--------------------------------------------------------------------
\subsection{Cross-Model Scaling Analysis}
\label{subsec:results_scaling}
%%--------------------------------------------------------------------

\begin{figure}[H]
    \centering
    \includegraphics[scale=0.8]{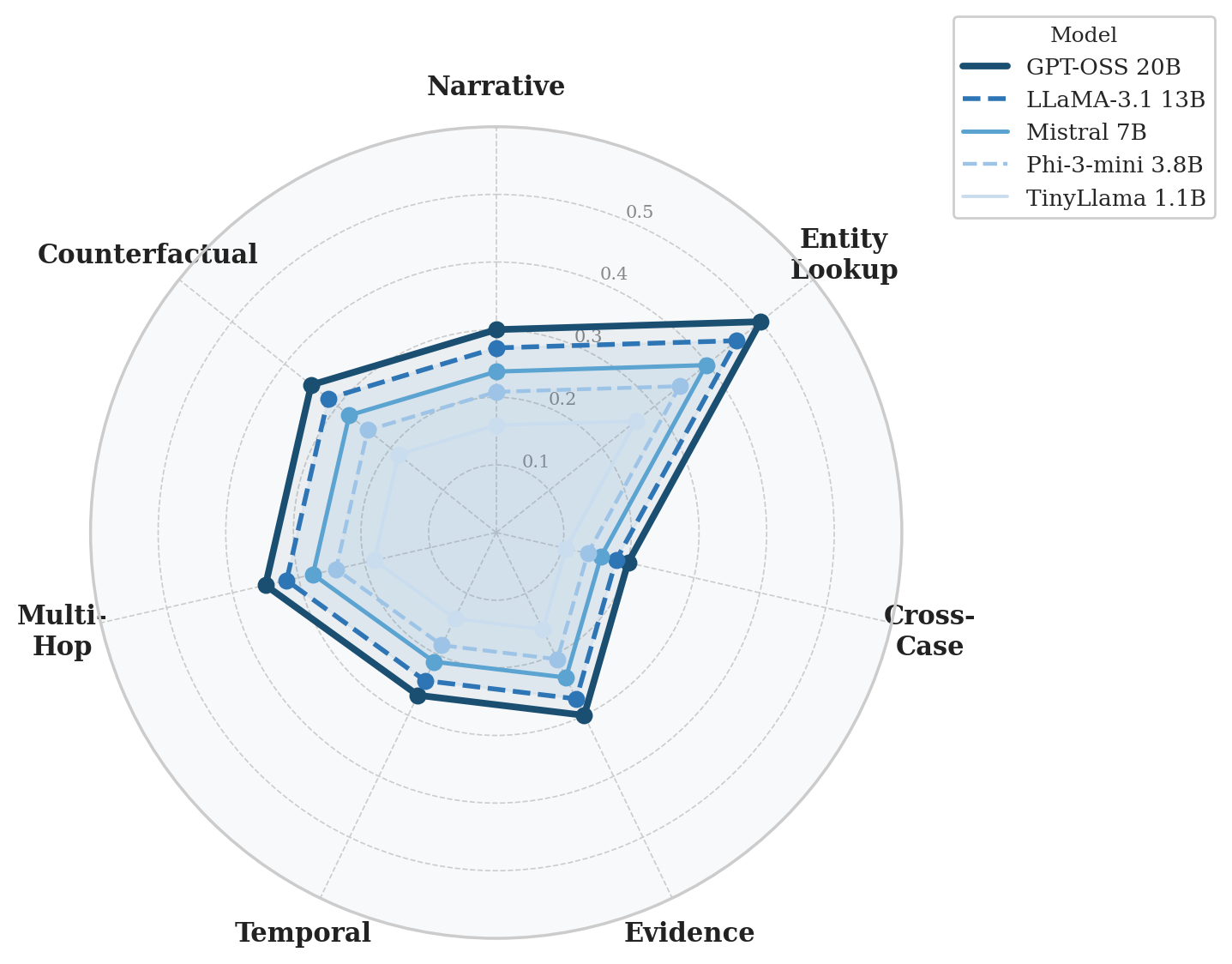}
    \caption{Radar Chart: Recall@5 by query type across five LLM models.}
    \label{fig:cross_type}
\end{figure}

Tables~\ref{tab:scaling} and Figure~\ref{fig:cross_type} show how TCAR-Gen
behaves across five generative models spanning an approximately
18-fold parameter range.

\begin{table}[htbp]
\centering
\caption{TCAR-Gen performance across five LLM models.
  \textbf{Bold}: best result per row.}
\label{tab:scaling}
\resizebox{\textwidth}{!}{%
\begin{tabular}{llccccc}
\toprule
\textbf{Metric} & \textbf{Stage} &
\textbf{GPT-OSS 20B} &
\textbf{LLaMA-3.1 13B} &
\textbf{Mistral 7B} &
\textbf{Phi-3-mini 3.8B} &
\textbf{TinyLlama 1.1B} \\
\midrule
\multicolumn{7}{l}{\textit{Retrieval Metrics}} \\
\midrule
Recall@3  & Retrieval & \best{0.1857} & 0.1689 & 0.1474 & 0.1289 & 0.0981 \\
Recall@5  & Retrieval & \best{0.3738} & 0.2945 & 0.2570 & 0.2247 & 0.1711 \\
Recall@10 & Retrieval & \best{0.5476} & 0.4981 & 0.4347 & 0.3800 & 0.2893 \\
NDCG@5    & Retrieval & \best{0.2354} & 0.2141 & 0.1869 & 0.1634 & 0.1244 \\
MRR       & Retrieval & \best{0.3357} & 0.3053 & 0.2665 & 0.2330 & 0.1773 \\
\midrule
\multicolumn{7}{l}{\textit{Generation Metrics}} \\
\midrule
Faithfulness      & Generation & \best{0.7872} & 0.2535 & 0.2212 & 0.1934 & 0.1472 \\
Answer Relevancy  & Generation & \best{0.8392} & 0.8143 & 0.7797 & 0.7471 & 0.6850 \\
Temporal Consist. & Generation & \best{1.0000} & 0.9703 & 0.9291 & 0.8903 & 0.8163 \\
\bottomrule
\end{tabular}%
}
\end{table}

Recall@5 decreases from $0.3738$ with GPT-OSS-20B to $0.1711$ with
TinyLlama-1.1B, an absolute reduction of $0.153$ across an 18-fold
decrease in parameter count. The reduction is modest relative to the model
size change. Retrieval quality in TCAR-Gen is partly structural, determined by the fixed context graph and hybrid scoring pipeline, and partly
model-conditioned, through the LLM-dependent query representation
$\mathbf{q}$ that drives the QCAP attention weights $\{\beta_v\}$.. The GNN encoder and
query-conditioned attention pooling also operate outside the language
model and determine chunk ranking directly. This separation matters in
practice because smaller models such as Phi-3-mini and TinyLlama still
retain useful retrieval coverage, with Recall@5 of $0.2247$ and $0.1711$
respectively. Generation quality depends much more strongly on model capacity.
Faithfulness decreases from $0.7872$ with GPT-OSS-20B to $0.2535$ with
LLaMA-3.1-13B and to $0.1472$ with TinyLlama-1.1B, a total reduction of
$0.640$. Smaller models therefore struggle less with evidence retrieval
than with synthesis of structured multi-step arguments. The evidence is
available, but the model cannot combine it reliably into explicit
claim-evidence relations.

Temporal Consistency decreases more gradually, from $1.0000$ to $0.8163$,
and remains above $0.8903$ for all models of at least 3.8B parameters.
The result suggests that the temporal penalty in Eq.~\eqref{eq:fusion}
provides useful structural support even for weaker models. Paths that
contain temporal violations are removed before synthesis, which reduces
the burden on the language model itself. Figure~\ref{fig:cross_type} shows that the ranking of query types by
Recall@5 remains stable across all five LLMs. Entity Lookup remains
the easiest type, whereas Cross-Case remains the most difficult. Query
difficulty in TCAR-Gen therefore, depends mainly on structural properties,
especially the number of reasoning hops and the degree of cross-document
bridging, rather than only on model size. Multi-Hop and Counterfactual
queries remain strong among the difficult categories at every scale. This
pattern reflects the effect of multi-hop context graph expansion, which
provides the same structural support across all models.

\section{Conclusion}\label{sec:conclusion}

This work presents a retrieval-augmented generation framework that integrates contextual, relational, and temporal reasoning within a single pipeline. The approach combines document-level context enrichment, query-conditioned graph construction, hybrid retrieval, and multi-branch reasoning. Context enrichment preserves document coherence, while the graph structure captures dependencies among entities and events. Hybrid retrieval uses semantic similarity, graph relevance, and temporal alignment to select evidence. Multi-branch reasoning evaluates alternative evidence paths and selects consistent outputs. The evaluation shows improvements in both retrieval and generation. The framework retrieves more relevant and complete evidence, especially for multi-hop and temporally constrained queries. Generated responses show higher faithfulness and better temporal consistency. The ablation results confirm that the graph structure and temporal reasoning are necessary for maintaining coherence. Future work can extend this framework in several directions. Larger and more diverse benchmarks can improve evaluation coverage. Adaptive graph construction and learned weighting can improve flexibility. Temporal modelling can be extended to capture richer event structures. Efficient reasoning strategies can reduce computational overhead. Integration with smaller models or distributed systems can improve scalability. 

%\bibliography{References}% common bib file

\begin{thebibliography}{10}
	
	\bibitem{qin2023chatgpt}
	C.~Qin, A.~Zhang, Z.~Zhang, J.~Chen, M.~Yasunaga, and D.~Yang, ``Is chatgpt a
	general-purpose natural language processing task solver?,'' {\em arXiv
		preprint arXiv:2302.06476}, 2023.
	
	\bibitem{peng2023check}
	B.~Peng, M.~Galley, P.~He, H.~Cheng, Y.~Xie, Y.~Hu, Q.~Huang, L.~Liden, Z.~Yu,
	W.~Chen, {\em et~al.}, ``Check your facts and try again: Improving large
	language models with external knowledge and automated feedback,'' {\em arXiv
		preprint arXiv:2302.12813}, 2023.
	
	\bibitem{borgeaud2022improving}
	S.~Borgeaud, A.~Mensch, J.~Hoffmann, T.~Cai, E.~Rutherford, K.~Millican, G.~B.
	Van Den~Driessche, J.-B. Lespiau, B.~Damoc, A.~Clark, {\em et~al.},
	``Improving language models by retrieving from trillions of tokens,'' in {\em
		International conference on machine learning}, pp.~2206--2240, PMLR, 2022.
	
	\bibitem{ram2023incontext}
	O.~Ram, Y.~Levine, I.~Dalmedigos, D.~Muhlgay, D.~Jannai, {\em et~al.},
	``In-context retrieval-augmented language models,'' {\em Transactions of the
		Association for Computational Linguistics}, vol.~11, pp.~1316--1331, 2023.
	
	\bibitem{lin2019kagnet}
	B.~Y. Lin, X.~Chen, J.~Chen, and X.~Ren, ``Kagnet: Knowledge-aware graph
	networks for commonsense reasoning,'' {\em arXiv preprint arXiv:1909.02151},
	2019.
	
	\bibitem{feng2020mhgrn}
	Y.~Feng, X.~Chen, B.~Y. Lin, P.~Wang, J.~Li, and X.~Ren, ``Scalable multi-hop
	relational reasoning for knowledge-aware question answering,'' in {\em
		Proceedings of the 2020 Conference on Empirical Methods in Natural Language
		Processing (EMNLP)}, pp.~1295--1309, 2020.
	
	\bibitem{li2023graphllm}
	Y.~Li, Z.~Li, P.~Wang, J.~Li, X.~Sun, H.~Cheng, and J.~X. Yu, ``A survey of
	graph meets large language model: Progress and future directions,'' {\em
		arXiv preprint arXiv:2311.12399}, 2023.
	
	\bibitem{lewis2020rag}
	P.~Lewis, E.~Perez, A.~Piktus, F.~Petroni, V.~Karpukhin, N.~Goyal,
	H.~K{\"u}ttler, M.~Lewis, W.-t. Yih, T.~Rockt{\"a}schel, S.~Riedel, and
	D.~Kiela, ``Retrieval-augmented generation for knowledge-intensive nlp
	tasks,'' in {\em Advances in Neural Information Processing Systems}, 2020.
	
	\bibitem{hu2025context}
	Y.~Hu, Q.~Li, D.~Zhang, J.~Yan, and Y.~Chen, ``Context-alignment: Activating
	and enhancing llm capabilities in time series,'' {\em arXiv preprint
		arXiv:2501.03747}, 2025.
	
	\bibitem{feng2025retrieval}
	J.~Feng, Q.~Wang, H.~Qiu, and L.~Liu, ``Retrieval in decoder benefits
	generative models for explainable complex question answering,'' {\em Neural
		Networks}, vol.~181, p.~106833, 2025.
	
	\bibitem{sun2018graftnet}
	H.~Sun, B.~Dhingra, M.~Zaheer, K.~Mazaitis, R.~Salakhutdinov, and W.~W. Cohen,
	``Open domain question answering using early fusion of knowledge bases and
	text,'' in {\em Proceedings of the 2018 Conference on Empirical Methods in
		Natural Language Processing}, pp.~4231--4242, 2018.
	
	\bibitem{sun2019pullnet}
	H.~Sun, T.~Bedrax-Weiss, and W.~W. Cohen, ``Pullnet: Open domain question
	answering with iterative retrieval on knowledge bases and text,'' in {\em
		Proceedings of the 2019 Conference on Empirical Methods in Natural Language
		Processing}, pp.~2380--2390, 2019.
	
	\bibitem{yang2024give}
	L.~Yang, H.~Chen, Z.~Li, X.~Ding, and X.~Wu, ``Give us the facts: Enhancing
	large language models with knowledge graphs for fact-aware language
	modeling,'' {\em IEEE Transactions on Knowledge and Data Engineering},
	vol.~36, no.~7, pp.~3091--3110, 2024.
	
	\bibitem{edge2024graphrag}
	D.~Edge, H.~Trinh, N.~Cheng, J.~Bradley, A.~Chao, A.~Mody, S.~Truitt,
	D.~Metropolitansky, R.~O. Ness, and J.~Larson, ``From local to global: A
	graph rag approach to query-focused summarization,'' {\em arXiv preprint
		arXiv:2404.16130}, 2024.
	
	\bibitem{he2024gretriever}
	X.~He, Y.~Tian, Y.~Sun, N.~V. Chawla, T.~Laurent, Y.~LeCun, X.~Bresson, and
	B.~Hooi, ``G-retriever: Retrieval-augmented generation for textual graph
	understanding and question answering,'' in {\em Advances in Neural
		Information Processing Systems}, 2024.
	
	\bibitem{linders2025knowledge}
	J.~Linders and J.~M. Tomczak, ``Knowledge graph-extended retrieval augmented
	generation for question answering,'' {\em Applied Intelligence}, vol.~55,
	no.~17, p.~1102, 2025.
	
	\bibitem{ding2025rdpg}
	L.~Ding, N.~Ding, Q.~Tao, {\em et~al.}, ``Enhancing graph multi-hop reasoning
	for question answering with {LLMs}: An approach based on adaptive path
	generation,'' {\em Journal of Intelligent Information Systems}, vol.~63,
	pp.~1455--1485, 2025.
	
	\bibitem{hussien2025rag}
	M.~M. Hussien, A.~N. Melo, A.~L. Ballardini, C.~S. Maldonado, R.~Izquierdo, and
	M.~{\'A}. Sotelo, ``Rag-based explainable prediction of road users behaviors
	for automated driving using knowledge graphs and large language models,''
	{\em Expert Systems with Applications}, vol.~265, p.~125914, 2025.
	
	\bibitem{li2025llm}
	L.~Li, Z.~Wang, J.~M. Jose, and X.~Ge, ``Llm supporting knowledge tracing
	leveraging global subject and student specific knowledge graphs,'' {\em
		Information Fusion}, p.~103577, 2025.
	
	\bibitem{chen2025temporal}
	K.~Chen, H.~Yu, Y.~Wang, X.~Song, X.~Zhao, Y.~Xie, L.~Gao, and A.~Li,
	``Temporal knowledge graph extrapolation with subgraph information
	bottleneck,'' {\em Expert Systems with Applications}, vol.~268, p.~126226,
	2025.
	
	\bibitem{yang2025gs}
	R.~Yang, J.~Zhu, J.~Man, H.~Liu, L.~Fang, and Y.~Zhou, ``Gs-kgc: A generative
	subgraph-based framework for knowledge graph completion with large language
	models,'' {\em Information Fusion}, vol.~117, p.~102868, 2025.
	
	\bibitem{xu2020tgat}
	D.~Xu, C.~Ruan, E.~Korpeoglu, S.~Kumar, and K.~Achan, ``Inductive
	representation learning on temporal graphs,'' {\em arXiv preprint
		arXiv:2002.07962}, 2020.
	
	\bibitem{bai2025few}
	L.~Bai, H.~Zhang, X.~An, and L.~Zhu, ``Few-shot multi-hop reasoning via
	reinforcement learning and path search strategy over temporal knowledge
	graphs,'' {\em Information Processing \& Management}, vol.~62, no.~3,
	p.~104001, 2025.
	
	\bibitem{liang2026spatiotemporal}
	X.~Liang, X.~Xu, R.~Ma, L.~Yan, and Z.~Ma, ``Spatiotemporal knowledge graph
	multi-hop reasoning based on large language models,'' {\em Engineering
		Applications of Artificial Intelligence}, vol.~164, p.~113229, 2026.
	
	\bibitem{zhang2024temporal}
	F.~Zhang, Z.~Zhang, F.~Zhuang, Y.~Zhao, D.~Wang, and H.~Zheng, ``Temporal
	knowledge graph reasoning with dynamic memory enhancement,'' {\em IEEE
		Transactions on Knowledge and Data Engineering}, vol.~36, no.~11,
	pp.~7115--7128, 2024.
	
	\bibitem{dong2024temporal}
	H.~Dong, P.~Wang, M.~Xiao, Z.~Ning, P.~Wang, and Y.~Zhou, ``Temporal inductive
	path neural network for temporal knowledge graph reasoning,'' {\em Artificial
		Intelligence}, vol.~329, p.~104085, 2024.
	
	\bibitem{meng2024multi}
	X.~Meng, L.~Bai, J.~Hu, and L.~Zhu, ``Multi-hop path reasoning over sparse
	temporal knowledge graphs based on path completion and reward shaping,'' {\em
		Information Processing \& Management}, vol.~61, no.~2, p.~103605, 2024.
	
	\bibitem{xu2025historical}
	W.~Xu, B.~Liu, M.~Peng, Z.~Jiang, X.~Jia, K.~Liu, L.~Liu, and M.~Peng,
	``Historical facts learning from long-short terms with language model for
	temporal knowledge graph reasoning,'' {\em Information Processing \&
		Management}, vol.~62, no.~3, p.~104047, 2025.
	
	\bibitem{dao2025hgct}
	H.~Dao, N.~Phan, T.~Le, and N.-T. Nguyen, ``{HGCT}: Enhancing temporal
	knowledge graph reasoning through extrapolated historical fact extraction,''
	{\em Knowledge-Based Systems}, vol.~316, p.~113358, 2025.
	
	\bibitem{zhu2024quadruple}
	L.~Zhu, W.~Zhao, and L.~Bai, ``Quadruple mention text-enhanced temporal
	knowledge graph reasoning,'' {\em Engineering Applications of Artificial
		Intelligence}, vol.~133, p.~108058, 2024.
	
	\bibitem{cai2025re}
	W.~Cai, M.~Li, X.~Shi, Y.~Fan, Q.~Zhu, and H.~Jin, ``Re-segnn: recurrent
	semantic evidence-aware graph neural network for temporal knowledge graph
	forecasting,'' {\em Science China information sciences}, vol.~68, no.~2,
	p.~122104, 2025.
	
	\bibitem{mei2024inductive}
	X.~Mei, L.~Yang, Z.~Jiang, X.~Cai, D.~Gao, J.~Han, and S.~Pan, ``An inductive
	reasoning model based on interpretable logical rules over temporal knowledge
	graph,'' {\em Neural Networks}, vol.~174, p.~106219, 2024.
	
	\bibitem{pan2025leveraging}
	Q.~Pan, L.~Yao, G.~Shen, X.~Han, Y.~Chen, and X.~Kong, ``Leveraging temporal
	validity of rules via llms for enhanced temporal knowledge graph reasoning,''
	{\em Knowledge-Based Systems}, p.~114094, 2025.
	
	\bibitem{ma2025historical}
	R.~Ma, L.~Wang, H.~Wu, B.~Gao, X.~Wang, and L.~Zhao, ``Historical trends and
	normalizing flow for one-shot temporal knowledge graph reasoning,'' {\em
		Expert Systems With Applications}, vol.~260, p.~125366, 2025.
	
	\bibitem{bai2025multi}
	L.~Bai, S.~Han, and L.~Zhu, ``Multi-hop interpretable meta learning for
	few-shot temporal knowledge graph completion,'' {\em Neural Networks},
	vol.~183, p.~106981, 2025.
	
	\bibitem{chen2025rule}
	T.~Chen, L.~Yang, Z.~Wang, and J.~Long, ``A rule-and query-guided reinforcement
	learning for extrapolation reasoning in temporal knowledge graphs,'' {\em
		Neural Networks}, vol.~185, p.~107186, 2025.
	
	\bibitem{li2025explainable}
	Q.~Li and G.~Wu, ``Explainable reasoning over temporal knowledge graphs by
	pre-trained language model,'' {\em Information Processing \& Management},
	vol.~62, no.~1, p.~103903, 2025.
	
	\bibitem{hu2025tempqa}
	Q.~Hu, X.~Tu, A.~Li, and B.~Yao, ``Tempqa: An llm-based framework for temporal
	knowledge graph question answering,'' {\em Knowledge-Based Systems},
	p.~114988, 2025.
	
	\bibitem{du2024serqa}
	C.~Du, X.~Li, and Z.~Li, ``Semantic-enhanced reasoning question answering over
	temporal knowledge graphs,'' {\em Journal of Intelligent Information
		Systems}, vol.~62, pp.~859--881, 2024.
	
	\bibitem{liu2025teqa}
	Q.~Liu, S.~Feng, and M.~Huang, ``{TEQA}: Temporal knowledge graph enhanced
	question answering,'' {\em Knowledge-Based Systems}, vol.~325, p.~113916,
	2025.
	
	\bibitem{wei2022cot}
	J.~Wei, X.~Wang, D.~Schuurmans, M.~Bosma, B.~Ichter, F.~Xia, E.~H. Chi, Q.~V.
	Le, and D.~Zhou, ``Chain-of-thought prompting elicits reasoning in large
	language models,'' {\em arXiv preprint arXiv:2201.11903}, 2022.
	
	\bibitem{wang2022selfconsistency}
	X.~Wang, J.~Wei, D.~Schuurmans, Q.~V. Le, E.~H. Chi, S.~Narang, A.~Chowdhery,
	and D.~Zhou, ``Self-consistency improves chain of thought reasoning in
	language models,'' {\em arXiv preprint arXiv:2203.11171}, 2022.
	
	\bibitem{reimers2019sentence}
	N.~Reimers and I.~Gurevych, ``Sentence-bert: Sentence embeddings using siamese
	bert-networks,'' in {\em Proceedings of the 2019 conference on empirical
		methods in natural language processing and the 9th international joint
		conference on natural language processing (EMNLP-IJCNLP)}, pp.~3982--3992,
	2019.
	
	\bibitem{es2024ragas}
	S.~Es, J.~James, L.~E. Anke, and S.~Schockaert, ``Ragas: Automated evaluation
	of retrieval augmented generation,'' in {\em Proceedings of the 18th
		conference of the european chapter of the association for computational
		linguistics: system demonstrations}, pp.~150--158, 2024.
	
	\bibitem{kazemi2019time2vec}
	S.~M. Kazemi and D.~Poole, ``Time2vec: Learning a vector representation of
	time,'' {\em arXiv preprint arXiv:1907.05321}, 2019.
	
\end{thebibliography}
\bibliographystyle{ieeetr}

%%%%%%%%%%%%%%%%%%

%%%%%%%%%%%%%%%%%%%

\end{document}